\documentclass{article}

\usepackage{PRIMEarxiv}

\usepackage[utf8]{inputenc} 
\usepackage[T1]{fontenc}    
\usepackage{hyperref}       
\usepackage{url}            
\usepackage{booktabs}       
\usepackage{amsmath,amsfonts}
\usepackage{nicefrac}       
\usepackage{microtype}      
\usepackage{lipsum}

\DeclareMathOperator*{\argmin}{arg\,min}
\usepackage{algorithmic}
\usepackage{algorithm}
\usepackage{array}
\usepackage{subfig}
\usepackage{textcomp}
\usepackage{stfloats}
\usepackage{url}
\usepackage{verbatim}
\usepackage{graphicx}
\usepackage{cite}
\usepackage{color}
\usepackage{multirow}
\usepackage{gensymb}
\usepackage{booktabs}
\usepackage{hyperref}
\usepackage{subfig}
\usepackage{fancyhdr}       
\usepackage{graphicx}       

\title{Causal Representation-Based Domain Generalization on Gaze Estimation
}

\author{
  Younghan Kim \\
  Gachon University, VisualCamp \\
  Seongnam, Korea\\
  \texttt{rio@visual.camp} \\
   \And
  Kangryun Moon \\
  Sungkyunkwan University \\
  Suwon, South Korea \\
  \texttt{kyle@visual.camp} \\
  \AND
  Yongjun Park, Yonggyu Kim \\
  VisualCamp \\
  Seoul, South Korea \\
  \texttt{\{cody, aiden\}@visual.camp} \\
}

\begin{document}
\maketitle

\begin{abstract}
The availability of extensive datasets containing gaze information for each subject has significantly enhanced gaze estimation accuracy. However, the discrepancy between domains severely affects a model's performance explicitly trained for a particular domain. In this paper, we propose the \emph{Causal Representation-Based Domain Generalization on Gaze Estimation (CauGE)} framework designed based on the general principle of causal mechanisms, which is consistent with the domain difference. We employ an adversarial training manner and an additional penalizing term to extract domain-invariant features. After extracting features, we position the attention layer to make features sufficient for inferring the actual gaze. By leveraging these modules, CauGE ensures that the neural networks learn from representations that meet the causal mechanisms' general principles. By this, CauGE generalizes across domains by extracting domain-invariant features, and spurious correlations cannot influence the model. Our method achieves state-of-the-art performance in the domain generalization on gaze estimation benchmark.
\end{abstract}


\section{Introduction} 

The human gaze is a crucial element of visual perception and communication as a fundamental means of conveying intentions, emotions, and attentiveness. By providing critical nonverbal cues, the gaze significantly influences social interactions. Its applications span various fields, including human-computer interaction~\cite{majaranta:eye}, virtual reality~\cite{soccini_gaze_2017}, and autonomous driving~\cite{pal_looking_2020}. One of the effective gaze estimation methods is the appearance-based approach, which utilizes a single image of the subject's face without relying on geometry information of the head and eyes. 

With the advent of deep learning techniques, Convolutional Neural Networks (CNNs) have demonstrated remarkable performance improvements in various computer vision tasks, and numerous gaze estimation methods using CNN have been proposed, including ~\cite{zhang_appearance-based_2015, krafka_eye_2016, zhang_its_2017}. The proliferation of large-scale datasets which have gaze information has also contributed significantly to advancing appearance-based gaze estimation~\cite{krafka_eye_2016, zhang_its_2017, funes_mora_eyediap_2014, kellnhofer_gaze360_2019, zhang_eth-xgaze_2020}.

\begin{figure}[t]
\centering
  \includegraphics[width=0.5\linewidth]{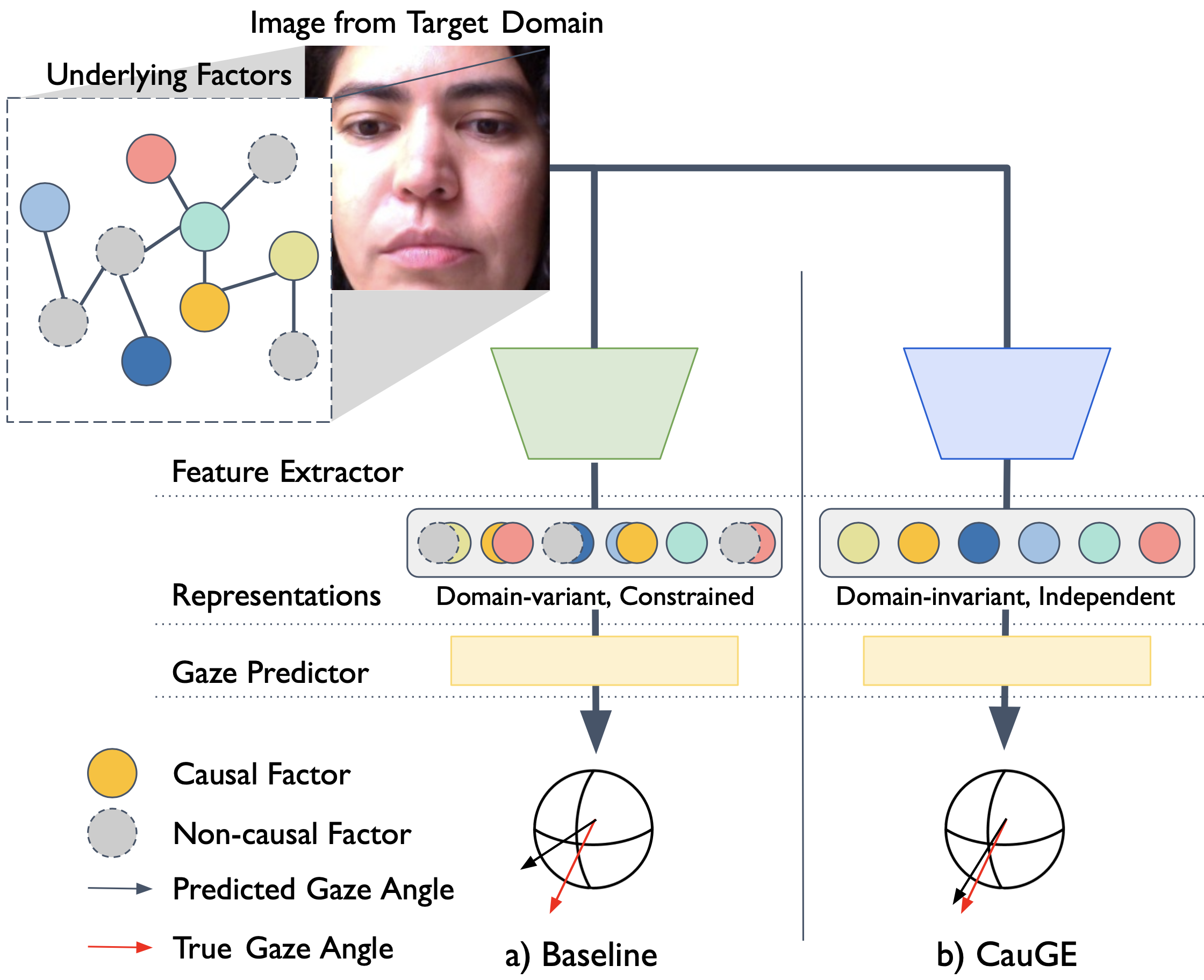}
  \caption{Dotted gray circles denote non-causal factors, and colored circles denote causal factors. a) Baseline, which is vanilla training, learns from representations mixed with non-causal factors and constrained with others. b) CauGE learns from representation separated from non-causal and independent factors. Learning such representation makes the model generalize well on unseen data.}
  \label{fig:causal_representation}
\end{figure}

Despite the progress made in improving performance within a dataset, a notable obstacle arises when applying neural networks trained on one dataset to another. We define this challenge resulted by two reasons. Firstly, a distribution gap between datasets may exist, characterized by differences such as subject appearance, image quality, shooting angle, illumination, and more. Machine learning models tend to suffer severe performance degradation when the probability distributions of the training data and testing data differ. While collecting all available data would ideally resolve this challenge, it is often an unfeasible undertaking. Secondly, estimation of the human gaze heavily relies on the eye region, which constitutes a relatively small portion of the face image~\cite{xu_learning_2023}. This problem makes the model vulnerable to spurious correlations. Consequently, a model may get influenced by non-causal factors, resulting in performance degradation on another dataset (See Figure~\ref{fig:causal_representation}(a)). Recent research has treated this issue as a domain adaptation problem~\cite{wang_generalizing_2019, wang_contrastive_2022}. However, domain adaptation methods typically require target dataset samples and additional optimization processes, making them less suitable for real-life scenarios and potentially degrading the user experience.

In this paper, we suggest a new approach to overcome these difficulties by developing a domain generalization method for gaze estimation, the ``Causal Representation-Based Domain Generalization on Gaze Estimation'' (CauGE) framework. Inspired by the concept of learning causal representations~\cite{lv_causality_2022}, our framework focuses on generating causal representations to extract domain-invariant features. By learning to generate these representations, the CauGE framework aims to improve generalization on unseen target domain datasets, thereby enhancing the robustness and performance of gaze estimation models (See Figure~\ref{fig:causal_representation}(b)). Also, as the CauGE framework focuses on the notion of causal mechanisms, this prevents overfitting to gaze-irrelevant features. By investigating the ideas above, we make the following contributions:

\begin{itemize}
  \item We propose the ``Causal Representation-Based Domain Generalization on Gaze Estimation (CauGE)'' method for domain generalization on gaze estimation. CauGE is an improved method to extract causal representations which meet general properties of causal factors on gaze estimation. Our research is the first to adapt the notion of causality in the gaze estimation field.
  \item Our method delivers superior results in the domain generalization of gaze estimation, outperforming state-of-the-art methods. Without access to the target dataset, our approach generalizes well on cross-dataset. Also, we conduct extensive experiments to assess our method's generalization capability.
\end{itemize}

\begin{figure*}
\centering
\subfloat[Common Cause Principle]{\label{fig:cm_a} \includegraphics[width=0.25\textwidth]{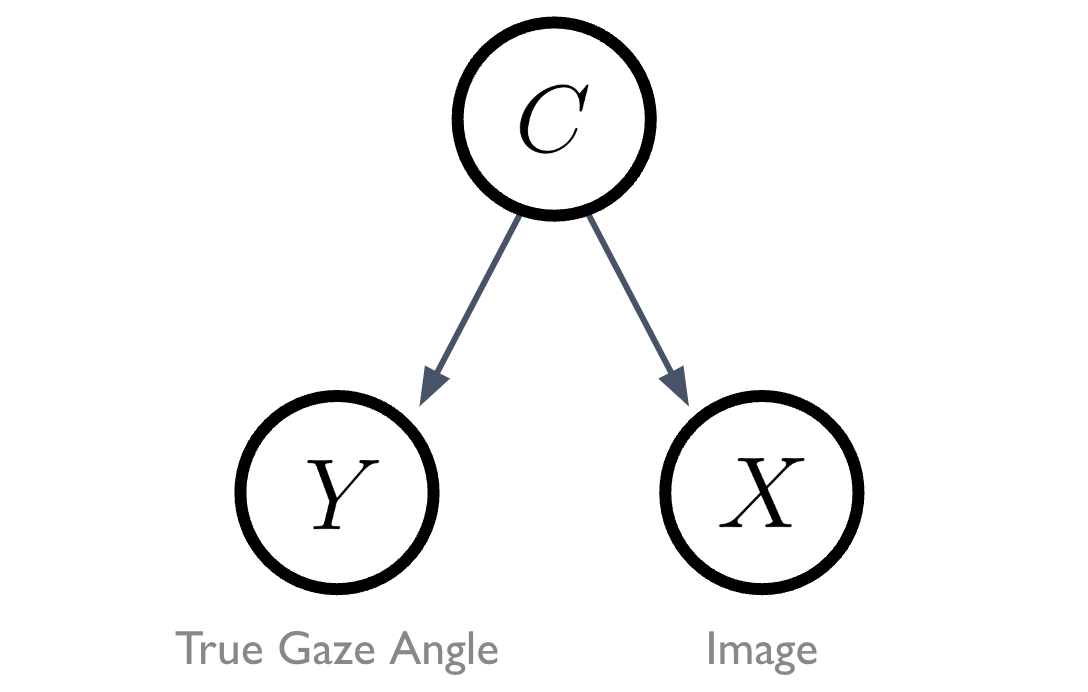}}%
\hfill
\subfloat[Stability]{\label{fig:cm_b} \includegraphics[width=0.23\textwidth]{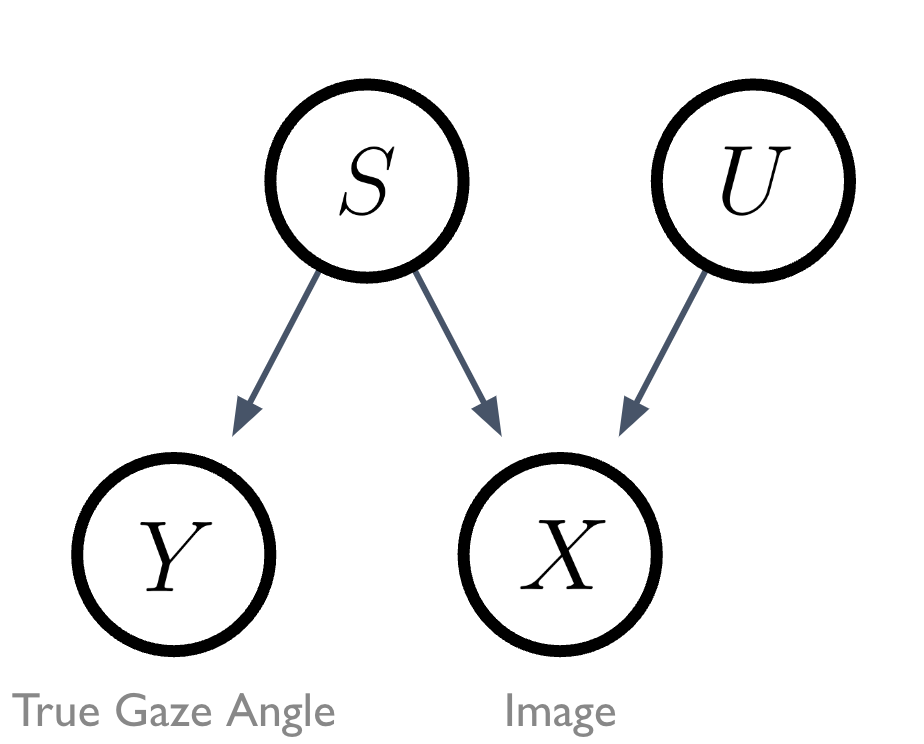}}%
\hfill
\subfloat[Modularity]{\label{fig:cm_c} \includegraphics[width=0.25\textwidth]{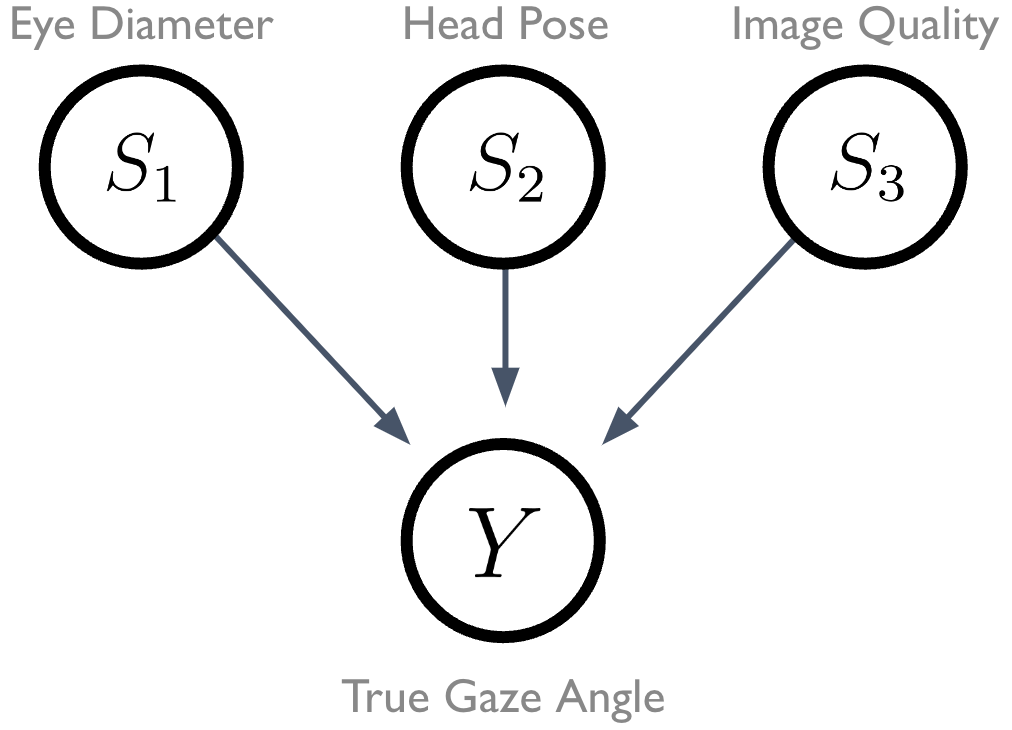}}%
\hfill
\subfloat[Causal Heterogeneity]{\label{fig:cm_d} \includegraphics[width=0.20\textwidth]{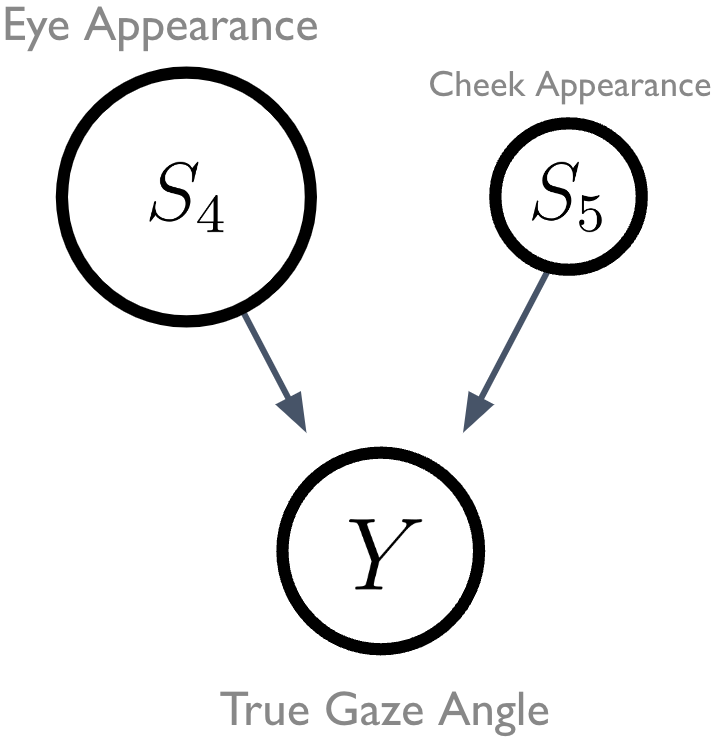}}%
\caption{Structural Causal Model, which describes the general principle of causal mechanisms. a) If $Y$ and $X$ are correlated, then $C$ should exist which is the common cause. b) When $S$ causes $Y$, this causal relationship is expected to be invariant. c) If we change one mechanism, it should not directly impact another mechanism. d) Not all causal factors are equal; some play a much more significant role than others. (The size of the circle describes the importance of a specific factor.)}
\label{fig:scm}
\end{figure*}
\section{Related Work}
\label{sec:related_work}
\subsection{Domain Generalization}
Domain adaptation (DA) and domain generalization (DG) are two distinct approaches to machine learning for handling the distribution gap. DA requires access to target domain data during training, while DG does not. Methods for DG include learning features that are invariant across domains~\cite{muandet_domain_2013}  and utilizing data augmentation. Augmenting the data help cover potential domains and includes traditional methods like image rotations and scaling, as well as more complex operations like style transfer~\cite{zhou_domain_2021} and Fourier transform-based methods~\cite{xu_fourier-based_2021}.

Recently, relying on causal mechanisms in DG has been proposed. The rationale behind leveraging causal mechanisms is that finding the causation of a specific event will be invariant to the domain, which leads to generalizing to unseen data. MatchDG \cite{mahajan_domain_2021} approximates to learn similar causal features by contrastive learning, assuming each class sample should share the same causal factor. Lv \textit{et al.} \cite{lv_causality_2022} defined three properties that causal factors should meet and tried to learn causal representation by imitating properties of causal factors. 

\subsection{Appearance-Based Gaze Estimation}

Appearance-based methods predict gaze directly from face images. Mainly CNN is widely used to extract features from face images \cite{zhang_appearance-based_2015, krafka_eye_2016,zhang_its_2017}. The advent of real-world gaze datasets~\cite{krafka_eye_2016, zhang_its_2017, funes_mora_eyediap_2014,  kellnhofer_gaze360_2019, zhang_eth-xgaze_2020}, made an appearance-based gaze approach which requires an extensive amount of data, become a more typical way to predict gaze directions. Zhang \textit{et al.} \cite{zhang_revisiting_2018} suggested a data normalization process for input images to reduce head pose variability and increase training efficiency. Recently, Cheng and Lu \cite{cheng_gazetr_2022} proposed an optimal variant of transformer architecture for gaze estimation. Despite these efforts, typical gaze estimation methods still have limitations of significantly decreasing accuracy on unseen domains.

\subsection{Cross-domain Gaze Estimation}

Cross-dataset method is a way to train a model on a particular dataset and evaluate the model on the other dataset with a different distribution from the training data. Typical gaze estimation models show performance degradation in the cross-dataset method due to domain shift. The DA method~\cite{wang_generalizing_2019, wang_contrastive_2022} was applied to gaze estimation fields to alleviate this weakness. However, DA approaches require target samples for the additional optimization process, which makes DA methods less suitable for real-world scenarios.

Researchers have also proposed DG works on gaze estimation. Cheng \textit{et al.} \cite{cheng_puregaze_2022} designed an adversarial training manner to extract gaze-relevant features while minimizing gaze-irrelevant features during training. Xu \textit{et al.} \cite{xu_learning_2023} disturbed the training data via adversarial attack or data augmentation and extracted the gaze-consistent features by aligning the gaze features. However, statistical dependence between image and label alone may not be sufficient to explain the underlying causal mechanisms, leading to instability across domains.

\section{Method} 

\subsection{General Principles of Causal Mechanisms for Adapting to Gaze Estimation}

In this section, we list out the general principle of causal mechanisms. We first explain Common Cause Principle (CCP)~\cite{reichenbach_direction_1991}. The CCP states that if two events are correlated, there must be a common cause responsible for this correlation, separate from the two events. Let us say we have $X$ for the image and $Y$ for the true gaze angle, which is correlated, i.e., $P(X \cap Y) \neq P(X)P(Y)$. According to the CCP, if $X$ and $Y$ are correlated, then there should exist some third event $C$ (the ``common cause'') that can account for this correlation (See Figure~\ref{fig:scm}(a)). There are two conditions for common cause $C$; First, $C$ should be correlated with both $X$ and $Y$, i.e., $P(X \vert C) \neq P(X), P(Y \vert C) \neq P(Y)$, and second, $X$ and $Y$ should be conditionally independent given $C$, i.e., $P(X \cap Y \vert C) = P(X \vert C)P(Y \vert C)$. Typical gaze estimation using machine learning methods estimates true gaze angle ($Y$) by modeling correlation from image ($X$). Modeling correlation between $X$ and $Y$ means that spurious correlation, which is not directly associated with true gaze angle, influences modeling. Therefore, by identifying $C$, we can know $Y$, the true gaze angle, by eliminating the spurious correlation between $X$ and $Y$. However, identifying $C$ between the image and true gaze angle is difficult since we do not have prior knowledge of which causes gaze angle. To alleviate this problem, inspired by Lv \textit{et al.} \cite{lv_causality_2022}, we aim to learn representations that meet the general principle of causal mechanisms: stability, modularity, and causal heterogeneity.

\begin{figure*}[t]
\centering
  \includegraphics[width=1\linewidth]{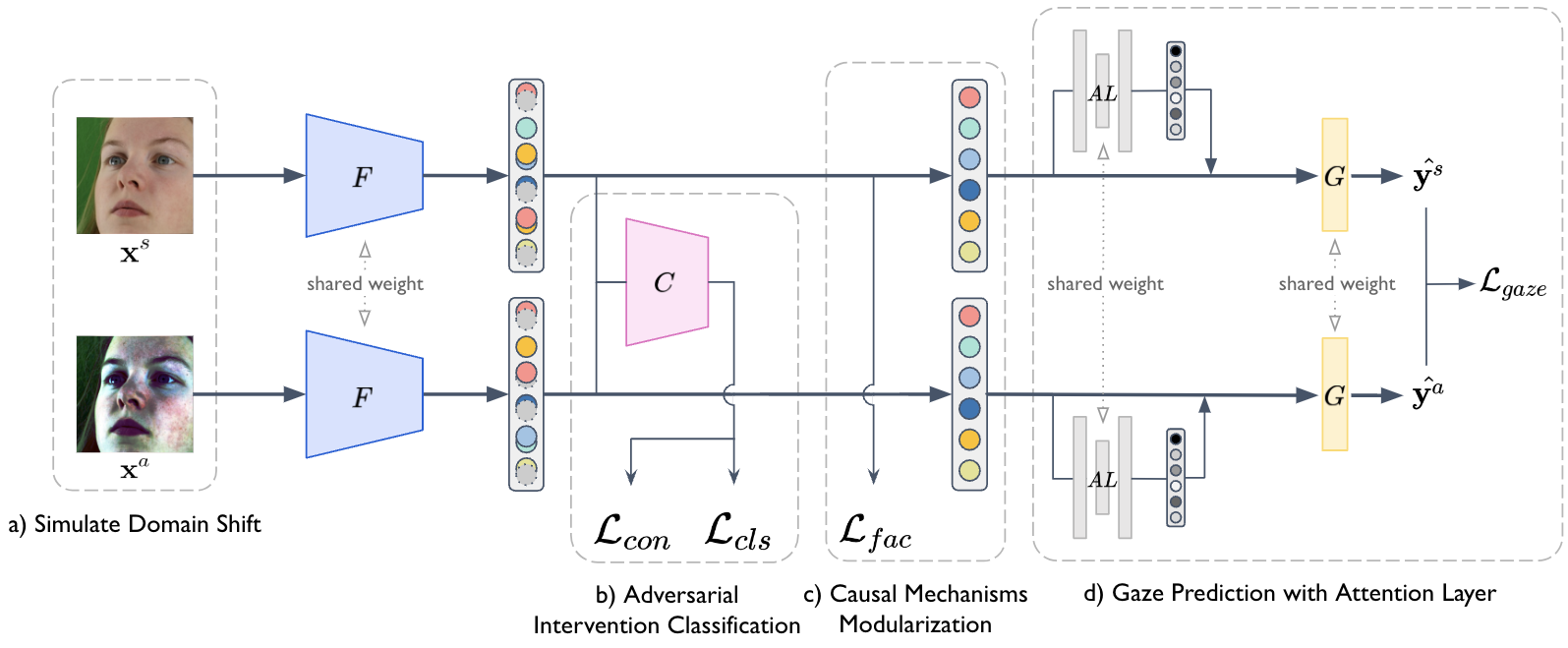}
  \caption{Overview of CauGE architecture. a) Simulate domain shift to capture the difference caused by the intervention. b) Force $F$ to extract domain-invariant features in adversarial intervention classification. c) Learn to extract independent causal representation with Factorization Loss. d) Strengthen representation with an attention layer and predict actual gaze angle with a gaze predictor.}
  \label{fig:overview}
\end{figure*}

\subsubsection{Stability}

Stability is a principle that suggests a true causal relationship should hold across different contexts or conditions, provided there is no interaction with another variable that is changing \cite{peters_elements_2017}. Only for an intuitive explanation, let us say one of the $S$, which causes both $X$ and $Y$  (See Figure~\ref{fig:scm}(b)), is the eyeball rotation of the subject in $X$, while one of the $U$, which only causes $X$, is illumination. Changing illumination will change the image. In contrast, it is also clear that changing illumination will not change the eyeball rotation of the subject if the subject's gaze is still. In other words, if we say ``$S$ causes $Y$'', this causal relationship is expected to be invariant, meaning it would remain consistent even when exposed to different situations or under different conditions.

\subsubsection{Modularity}

The principle of modularity suggests that each causal mechanism operates independently \cite{pearl_causality_2009}. The modularity principle means that changing one mechanism should not directly impact another. For instance, let us consider a system where multiple variables interact. Imagine a gaze estimation system where the eye diameter of the subject $S_1$, head pose $S_2$, and image quality $S_3$ all affect outcome $Y$, the true gaze angle. The mechanism by which $S_1$ affects $Y$ is independent of the mechanism by which $S_2$ affects $Y$, and both are independent of the mechanism by which $S_3$ affects $Y$ (See Figure \ref{fig:scm}(c)). If we were to intervene and change the relationship between $S_1$ and $Y$, it would not impact the relationships between $S_2$ and $Y$ or $S_3$ and $Y$. The modularity assumption makes our models more robust to environmental changes or distributional shifts. If the mechanisms are independent, a change in one part of the environment, such as changed head pose or widened eye, should only affect the mechanism that directly depends on it and not others.

\subsubsection{Causal Heterogeneity}


Causal Heterogeneity is a fundamental concept in the field of causal inference that highlights the variability in the effects of a treatment or intervention across different segments of a population. This concept acknowledges that the impact of any given cause is not universally consistent but can vary significantly based on individual characteristics, contextual factors, or other relevant variables. For instance, both appearance of the subject's eye $S_4$ and cheek $S_5$ matter to the true gaze angle. However, the appearance of the subject's eye matters more than the cheek (See Figure~\ref{fig:scm}(d)). Causal Heterogeneity is a vital assumption for fitting causal models to data, as it allows us to infer the presence of a causal link from an observed association.

\subsection{The CauGE Framework}

The CauGE Framework (See Figure~\ref{fig:overview}) ensures that these representations meet the general properties of causal factors, as listed before. We simulate domain shift (Figure~\ref{fig:overview}(a)) by manually intervening upon the input image $\mathbf{x}$ purposely to capture the difference caused by the intervention to effectively separate causal factors $S$ from input image $\mathbf{x}$ which is mixed with non-causal factors $U$. The framework extracts $S$, which meets the stability principle by leveraging adversarial intervention classification task (Figure~\ref{fig:overview}(b)). We aim to establish independence within the representation to meet the modularity principle for each causal mechanism (Figure~\ref{fig:overview}(c)). To strengthen representation for the gaze estimation task to meet the causal heterogeneity principle, we utilize the attention layer and predict gaze angle (Figure~\ref{fig:overview}(d)).

For a given source domain dataset $\mathcal{D}_s = \{(\mathbf{x}^{s}, \mathbf{y}^{s})\}$, we sample a single batch data, defined as $(\mathbf{x}^{s}, \mathbf{y}^{s})$, which is input image and its corresponding gaze angle, respectively. A set of augmentation techniques $A(\cdot)$ is applied in $\mathbf{x}^{s}$, formulated as $\mathbf{x}^{a} = A(\mathbf{x}^{s})$. The feature extractor ${F}$ which has $\mathbf{x}^{s}$ and $\mathbf{x}^{a}$ as input, sequentially. These representations $\mathbf{z}^{s} = {F}(\mathbf{x}^{s}), \space \mathbf{z}^{a} = {F}(\mathbf{x}^{a})$ are fed into the intervention classifier ${C}$. Simultaneously,  $\mathbf{z}^{s}$ and $\mathbf{z}^{a}$ go through attention layer ${AL}$: $\mathbf{w}^{s} = {AL}(\mathbf{z}^{s}), \mathbf{w}^{a} = {AL}(\mathbf{z}^{a})$. Finally, a gaze predictor ${G}$ has $\mathbf{z}^{s} \odot \mathbf{w}^{s}$ and $\mathbf{z}^{a} \odot \mathbf{w}^{a}$ to predict actual gaze angle: $\hat{\mathbf{y}}^{s} = {G}(\mathbf{z}^{s} \odot \mathbf{w}^{s})$, $\hat{\mathbf{y}}^{a} = {G}(\mathbf{z}^{a} \odot \mathbf{w}^{a}).$

\subsubsection{Simulating Domain Shift}

In order to distinguish $S$ from $U$ in an input image $\mathbf{x}$, we intentionally manipulate the image to create a domain shift and observe any differences caused by the intervention. $S$ should remain stable across different contexts or conditions, which is $U$, i.e., \begin{equation} \label{eq_domain_invariant}
P(Y \vert S, do(U = u_{i})) = P(Y \vert S, do(U = u_{j})) \forall  u_{i}, u_{j} \in U,
\end{equation}
\noindent~where $do(\cdot)$ refers to intervention~\cite{pearl_causality_2009}.

To simulate intervention, we use AugMix~\cite{hendrycks_augmix_2020}. AugMix is a data augmentation technique that enhances models' robustness. AugMix mixes the results of chain augmentations in convex combinations. By combining augmentations, the increase in diversity may result in the sample being outside of the data manifold. The authors suggest that their method of combining augmentations produces realistic transformations. We remove sheerX, sheerY, translate and rotate operations to prevent the change of gaze angle.

\subsubsection{Adversarial Intervention Classification}

After simulating domain shift, we force $F$ to extract domain-invariant features. As performing an intervention on $U$ does not change $S$, the representation generated from the original and augmentation-applied images should be identical (Equation~\ref{eq_domain_invariant}). We design an adversarial training manner to distinguish $S$ from $U$ to achieve this. This process engages two neural networks in a competitive scenario: $F$, which transforms $\mathbf{x}^{s}$ and $\mathbf{x}^{a}$ into a new feature representation, and $C$, tasked with distinguishing whether the image was intervened or not from $\mathbf{z}^{s}$ and $\mathbf{z}^{a}$. $F$ progressively learns to generate representations that confound $C$ through this adversarial process, thereby achieving domain invariance.

$C$ has representation generated from original and augmentation-applied images, $\mathbf{z}^s$, $\mathbf{z}^a$ as input. The objective of $C$ is to classify whether the representation from the input image is original or intervened: \begin{equation} \label{intervention_classifier_loss}
\begin{aligned}
&\mathcal{L}_{cls} = {-} \mathbb{E}_{\mathbf{z}^{s} \sim p_{\mathbf{z}^{s}}(\mathbf{z}^{s})} \big[ \log C(\mathbf{z}^{s}) \big] \\
&\hspace{1.08cm} {-}\mathbb{E}_{\mathbf{z}^{a} \hspace{-0.02cm} \sim p_{\mathbf{z}^{a}\hspace{-0.02cm}}(\mathbf{z}^{a}\hspace{-0.02cm})} \big[ \log (1 - C(\mathbf{z}^{a})) \big],
\end{aligned}
\end{equation} where the expected output of $C$ for original images and augmentation-applied is 1, 0, respectively.

To make the representations from $F$ not change by intervention, we use the inverted label~\cite{tzeng_adversarial_2017}. This misclassification of $C$ caused by the inverted label exerts pressure on the $F$ during the adversarial training process. $F$, aiming to minimize the intervention confusion loss:\begin{equation} \label{intervention_confusion_loss}
\mathcal{L}_{con} = - \mathbb{E}_{\mathbf{z}^{a} \sim p_{\mathbf{z}^{a}}(\mathbf{z}^{a})} \big[ \log C(\mathbf{z}^{a}) \big],
\end{equation} is thus compelled to generate features beneficial for the primary task and capable of confounding the $C$. Essentially, the $F$ learns to generate representations that resemble those from the target when processing source domain data and vice versa, thus leading to generating domain-agnostic representations.

\subsubsection{Causal Mechanisms Modularization}

We aim to ensure that any two dimensions in the representations are not dependent on one another. Suppose $S$ is a set, i.e., $S = \{s_1, s_2, s_3, \dots, s_N\}$. $P(s_{i} | PA_{i})$, where $PA_{i}$ denotes the parent node of $s_i$, within $S$ should not affect any other mechanisms $P(s_{j} | PA_{j})$. Additionally, knowing certain mechanisms $P(s_{i} | PA_{i})$ should not provide insight into other mechanisms $P(s_{j} | PA_{j})$.

We utilize the factorization loss from Lv \textit{et al.} \cite{lv_causality_2022}. Under the assumption that $\mathbf{z}^{s}$ and $\mathbf{z}^{a}$ are domain-invariant, correlation matrix $M$ is formulated as: \begin{equation} \label{correlation_matrix}
M_{ij} = \frac{\langle \mathbf{z}^{s}_{i}, \mathbf{z}^{a}_{j} \rangle}{\lvert \lvert \mathbf{z}^{s}_{i} \rvert  \rvert \, \lvert \lvert \mathbf{z}^{a}_{j} \rvert \rvert}, \hspace{0.05cm} i, j \in 1, 2, \cdots, N,
\end{equation}
\noindent where $i, j$ indicates specific dimension of $\mathbf{z}^s$ and $\mathbf{z}^a$. Please note that the $\mathbf{z}^{s}$ and $\mathbf{z}^{a}$ is normalized with Z-score normalization for forming $M$.

We want to maximize the correlation between the identical dimension of $\mathbf{z}^s$ and $\mathbf{z}^a$ while minimizing the correlation between the two dimensions. Final factorization loss is: \begin{equation} \label{factorization_loss}
\mathcal{L}_{fac} = \frac{1}{2} \cdot \lvert \lvert M - I \rvert \rvert^{2}_{F}.
\end{equation}

\subsubsection{Gaze Prediction with Attention Layer}

Rather than predicting the gaze angle from the original representation, we adopt soft attention from SE-Net~\cite{hu_squeeze-and-excitation_2018} to weigh the causal factors to make $S$ meet the causal heterogeneity principle. By generating a channel-wise attention map and rescaling the representation and the attention map, we give information about which causal factor matters more. The process consists of two main steps: Squeeze and Excitation.

The Squeeze step combines spatial information from each channel to generate channel-wise statistics. The Squeeze step uses global average pooling across the height and width dimensions, resulting in a single scalar for each channel. The Excitation step utilizes these statistics to capture channel-wise dependencies through a gating mechanism that uses a sigmoid function and generates a channel-wise attention map that ranges from 0 to 1. The networks rescale their representations using a channel-wise attention map, focusing more on valuable features and suppressing less useful ones. This process enables the network to recalibrate its feature responses in a channel-wise manner, leading to an improvement in its representational power.

After going through the attention layer, we feed these representations to the $G$. $G$ predicts the actual gaze angle, and to optimize based on the prediction, we calculate L1 Loss for each prediction and, i.e., \begin{equation} \label{gaze_loss}
\mathcal{L}_{gaze} = \frac{1}{2} \cdot \lvert \lvert \mathbf{y}^s - \hat{\mathbf{y}}^s\rvert \rvert_{1} + \frac{1}{2} \cdot \lvert \lvert \mathbf{y}^s - \hat{\mathbf{y}}^a \rvert \rvert_{1}.
\end{equation}

\subsection{Training and Inference}

We formulate the primary objective of the CauGE framework as follows: 
\begin{equation} \label{l_prim}
  \begin{aligned}
    & \mathcal{L}_{prim}(F, AL, G) = \lambda_{con} \cdot \mathcal{L}_{con}(F) \\ 
    & \hspace{2.51cm} + \lambda_{fac}\cdot \mathcal{L}_{fac}(F) \\
    & \hspace{2.51cm} + \lambda_{gaze} \cdot \mathcal{L}_{gaze}(F, AL, G)
  \end{aligned}
\end{equation}

\noindent where $\lambda_{con}, \lambda_{fac}$ and $\lambda_{gaze}$ denote the coefficient for each loss function. We empirically set these hyper-parameters as 3, 2, and 5, respectively. Our aim of the framework is formulated as:\begin{equation} \label{full_objective}
  \begin{aligned}
    & \argmin_{F, AL, G} \mathcal{L}_{prim}(F, AL, G), \\
    & \argmin_{C} \hspace{0.05cm} \mathcal{L}_{cls}(C). \\
  \end{aligned}
\end{equation} By minimizing $\mathcal{L}_{prim}$ and training in adversarial manner with $\mathcal{L}_{cls}$, CauGE framework can extract features that meets the general principles of causal mechanisms which achieve domain invariance.

At inference phase, we use target dataset $\mathcal{D}_t = \{(\mathbf{x}^{t}, \mathbf{y}^{t})\}$. As our targeting task is DG, we restrict updating model parameters using $\mathcal{D}_t$. We remove applying augmentation to $\mathbf{x}^t$ and $C$. We formulate the inference phase as follows: \begin{equation} \label{inference} \mathbf{z}^t = F(\mathbf{x}^t), \hspace{0.05cm} \mathbf{w}^{t} = {AL}(\mathbf{z}^{t}), \hspace{0.05cm} \hat{\mathbf{y}}^t = G(\mathbf{z}^{t} \odot \mathbf{w}^{t}).
\end{equation}

\section{Experiments}
\label{sec:experiment}

\subsection{Experimental Setup}
\subsubsection{Datasets}
For our method assessment, we utilized Gaze360 ($D_G$)~\cite{kellnhofer_gaze360_2019} and ETH-XGaze ($D_E$)~\cite{zhang_eth-xgaze_2020} as source domain datasets due to their diverse range of subjects, head poses, and gaze angles. For target domain datasets, we employed MPIIFaceGaze ($D_M$)~\cite{zhang_its_2017} and EYEDIAP ($D_D$)~\cite{funes_mora_eyediap_2014}. We performed experiments on four DG tasks, $D_E$ $\xrightarrow{}$ $D_M$, $D_E$ $\xrightarrow{}$ $D_D$, $D_G$ $\xrightarrow{}$ $D_M$, and~$D_G$~$\xrightarrow{}$~$D_D$.

For a fair comparison with previous DG methods on gaze estimation, we follow the same data preprocessing with \cite{cheng_puregaze_2022}. $D_E$ contains 80 subjects in the training set, and we randomly select five subjects and exclude them from the training set for validation. For $D_G$, we only include images when the face of the subject is visible in the image. We follow the standard protocol for $D_M$. For $D_D$, we follow the steps in \cite{zhang_its_2017}. For rectifying the data for $D_E$, $D_M$, and $D_D$, we employ the method from Zhang \textit{et al.} \cite{zhang_revisiting_2018}. As the authors preprocess $D_G$, we use the data which the authors provide.

\subsubsection{Implementation Details}

The input to the model is a 224x224 RGB image normalized using ImageNet's mean and standard deviation. We use ResNet-18~\cite{he_deep_2016} as the backbone network, with the last layer replaced with an MLP layer for gaze estimation. We set the reduction ratio of the attention layer as 8. We trained our model using the Adam optimizer with a learning rate $10^{-3}$ and set $\beta_1 = 0.9$ and $\beta_2 = 0.95$. We set the batch size as 512 and trained the model for 50 epochs. The experiments were performed on a single NVIDIA A100 GPU with 40 GB of memory.

\subsection{Comparison with Existing Methods}

In this part, we have conducted an empirical study to compare the effectiveness of our suggested approach with the current state-of-the-art methods. We use the ResNet-18 with a single MLP Layer for the baseline model. Training for the baseline model involves calculating the L1 loss between the predicted gaze direction and its corresponding ground truth. We use the results in PureGaze~\cite{cheng_puregaze_2022} and GcFE~\cite{xu_learning_2023} to compare DG techniques. We refer to GcFE as the feature extraction component of their approach. As shown in Table~\ref{tab:naive_compare}, our method surpasses the previous approaches of gaze estimation on cross-dataset evaluation. Our method has scored decreased angular error by 1.37 and 0.71 with PureGaze and GcFE, on average. As the results indicate, our proposed method has proven effective in achieving DG.

The baseline results in this study differ from those reported in other research papers. This discrepancy can be attributed to variations in training specifics and minor modifications in the baseline architecture. However, it is important to note that the proposed CauGE model and the baseline were trained under identical conditions, including the train/validation split for the ETH-XGaze dataset and hyperparameter settings. The primary objective of comparing CauGE with the baseline is to assess how well the model generalizes across different domains by solely leveraging the CauGE framework, without altering any other aspects. Despite the differences in baseline results compared to other works, this comparison is considered valid and informative because the training environment remains consistent. By maintaining this consistency, the experiment aims to isolate and evaluate the impact of the CauGE framework on the model’s ability to generalize across domains effectively.

\begin{table}[t!]
\centering
\caption{Comparison with Existing Method on Gaze Estimation. The angular error refers to the degree of the angular measuring unit. We indicate the best result in each column in bold text and the second best in underlined text.}
\resizebox{0.5\columnwidth}{!}{%
\begin{tabular}{@{}llllll@{}}
\toprule
\multicolumn{1}{c}{Method} &
  \begin{tabular}[c]{@{}l@{}}$D_{E}$ \\ $\xrightarrow{}$ $D_{M}$\end{tabular} &
  \begin{tabular}[c]{@{}l@{}}$D_{E}$ \\ $\xrightarrow{}$ $D_{D}$\end{tabular} &
  \begin{tabular}[c]{@{}l@{}}$D_{G}$ \\ $\xrightarrow{}$ $D_{M}$\end{tabular} &
  \begin{tabular}[c]{@{}l@{}}$D_{G}$\\ $\xrightarrow{}$ $D_{D}$\end{tabular} &
  \multicolumn{1}{c}{Avg.} \\ \midrule
Baseline    & 11.37 & 20.92 & 11.63 & 9.52  & 13.36   \\ 
PureGaze    & 7.08  & 7.48  & 9.28  & 9.32  & 8.29   \\
GcFE        & \underline{6.50}  & \underline{7.44}  & \underline{7.55}  & \underline{9.03}  & \underline{7.63}   \\ 
CauGE(ours) & \textbf{6.28}  & \textbf{7.39}  & \textbf{6.36} & \textbf{7.65} & \textbf{6.92}   \\ \bottomrule
\end{tabular}%
}

\label{tab:naive_compare}
\end{table}

\begin{table}[t!]
\centering
\caption{Ablation studies of CauGE. Experiments indicate that intervention confusion loss (Intr.), factorization loss (Fact.), and attention layer module (Attn.) complement each other, and all are necessary.}
\resizebox{0.5\columnwidth}{!}{%
\begin{tabular}{@{}ccclllll@{}}
\toprule
\multicolumn{1}{c}{Intr.} &
\multicolumn{1}{c}{Fact.} &
\multicolumn{1}{c}{Attn.} &
  \begin{tabular}[c]{@{}l@{}}$D_{E}$ \\ $\xrightarrow{}$ $D_{M}$\end{tabular} &
  \begin{tabular}[c]{@{}l@{}}$D_{E}$ \\ $\xrightarrow{}$ $D_{D}$\end{tabular} &
  \begin{tabular}[c]{@{}l@{}}$D_{G}$ \\ $\xrightarrow{}$ $D_{M}$\end{tabular} &
  \begin{tabular}[c]{@{}l@{}}$D_{G}$\\ $\xrightarrow{}$ $D_{D}$\end{tabular} &
  \multicolumn{1}{c}{Avg.} \\ \midrule
- & -       & -       &  11.37 & 20.92 & 11.63 & 9.52  & 13.36 \\ \midrule
$\checkmark$ & -       & -       & 7.32 & 8.10 & 6.81 & 8.04 & 7.56 \\
-       & $\checkmark$ & -       & 7.05 & 8.32 & 6.37 & 8.20 & 7.48 \\
- & - &   $\checkmark$    & 7.31 & 7.49 & 6.88 & 8.51 & 7.54 \\ \midrule
$\checkmark$ & $\checkmark$ & -       & 7.67 & 9.52 & 8.53 & 8.34 & 8.51\\
$\checkmark$ & - & $\checkmark$       & 6.88 & 8.43 & 6.73 & 8.58 & 7.65 \\
- & $\checkmark$ & $\checkmark$       & 7.13 & 7.43 & 6.90 & 7.75 & 7.30 \\ \midrule
$\checkmark$ & $\checkmark$ & $\checkmark$ & \textbf{6.28} & \textbf{7.39} & \textbf{6.36} & \textbf{7.65} & \textbf{6.92 }\\ \bottomrule
\end{tabular}
}

\label{tab:as}
\end{table}

\subsection{Additional Experiments}

\subsubsection{Ablation Studies}

In this section, we delve into the contribution of the intervention confusion loss, factorization loss, and attention layer module in the CauGE framework. Table~\ref{tab:as} presents the results of various versions of CauGE, which demonstrate that all versions surpass the baseline. Moreover, factorization loss, alone or in combination with the attention layer, yields better results than intervention confusion loss in the same conditions. This result is due to the fact that factorization loss penalizes the feature extractor for extracting independent and domain-invariant features. In contrast, intervention confusion loss only penalizes the feature extractor for invariance to the domain. However, by implementing an adversarial training approach with the domain classifier, the penalizing term becomes more reliable as the domain classifier computes the non-linear relationship between two representations.

One experiment demonstrates that the absence of an attention layer in conjunction with intervention confusion loss and factorization loss leads to the poorest outcomes. This experiment emphasizes the significance of the attention layer and the necessity of weighing causal mechanisms to attain optimal results during the inference phase. Ultimately, CauGE framework, which incorporates all three components, surpasses the other models in the benchmark, indicating that the three modules mutually reinforce each other and that each of them is integral to achieving precise gaze estimation.

\subsubsection{Domain Shift Simulation Competence}
Previous DG work leverages Fourier transform~\cite{xu_fourier-based_2021} to split an image into amplitude and phase information. Xu \textit{et al.} \cite{xu_fourier-based_2021} argued manipulating amplitude information can simulate domain shift while being consistent with task-relevant information, such as object class in the classification task. To compare the competency of simulating domain shift for gaze estimation between AugMix in CauGE and Fourier transform-based data augmentation, we remove AugMix and adapt Fourier transform-based data augmentation. In order to effectively augment the amplitude information of an image, we employ a linear blending technique. This involves combining the original image's amplitude information with that of a randomly selected image from the same source domain. The results (See Table~\ref{tab:variants}) show that leveraging AugMix is more effective in generalizing on unseen data. This result suggests that AugMix can diversify data samples without losing gaze information in the image.

\begin{table}[t!]
\centering
\caption{Comparison with Fourier transform-based augmentation (FT-Aug) and adversarial mask module (AMM). Angular errors in degrees
are shown. The results show that CauGE outperforms these variants.}
\begin{tabular}{@{}llllll@{}}
\toprule
\multicolumn{1}{c}{Method} &
  \begin{tabular}[c]{@{}l@{}}$D_{E}$ \\ $\xrightarrow{}$ $D_{M}$\end{tabular} &
  \begin{tabular}[c]{@{}l@{}}$D_{E}$ \\ $\xrightarrow{}$ $D_{D}$\end{tabular} &
  \begin{tabular}[c]{@{}l@{}}$D_{G}$ \\ $\xrightarrow{}$ $D_{M}$\end{tabular} &
  \begin{tabular}[c]{@{}l@{}}$D_{G}$\\ $\xrightarrow{}$ $D_{D}$\end{tabular} &
  \multicolumn{1}{c}{Avg.} \\ \midrule
FT-Aug  & 7.35 & 7.47 & 7.30 & 10.15  &  8.06  \\
AMM    & 10.67  & 12.74  & 10.88  & 11.95  & 11.56   \\
CauGE & 6.28  & 7.39  & 6.36  & 7.65  & 6.92   \\ \bottomrule
\end{tabular}%

\label{tab:variants}
\end{table}

\subsubsection{Effectiveness of Attention Layer}
One of the differences between our method and Lv \textit{et al.} \cite{lv_causality_2022} is the module before the gaze predictor. Lv \textit{et al.} \cite{lv_causality_2022} proposed the adversarial mask module, which identifies the dimensions that have less impact on the primary task and compels them to contribute more. We experiment to compare our attention layer and adversarial mask module on the CauGE framework. To utilize the adversarial mask module in our framework, we add a gaze predictor and follow the implementation. The goal of the adversarial mask module is to involve more underlying causal factors and squeeze out non-causal factors, while our attention layer provides information on the importance of certain causal factors. The result (See Table~\ref{tab:variants}) shows that the attention layer is better at generalizing unseen data than the adversarial mask module. This experiment suggests that weighing a specific factor is essential in learning causal representation.

\subsection{Visual Report}

\begin{figure}[t!]
\centering
\subfloat[Baseline]{\label{fig:illum_cam_baseline} \includegraphics[width=0.49\linewidth]{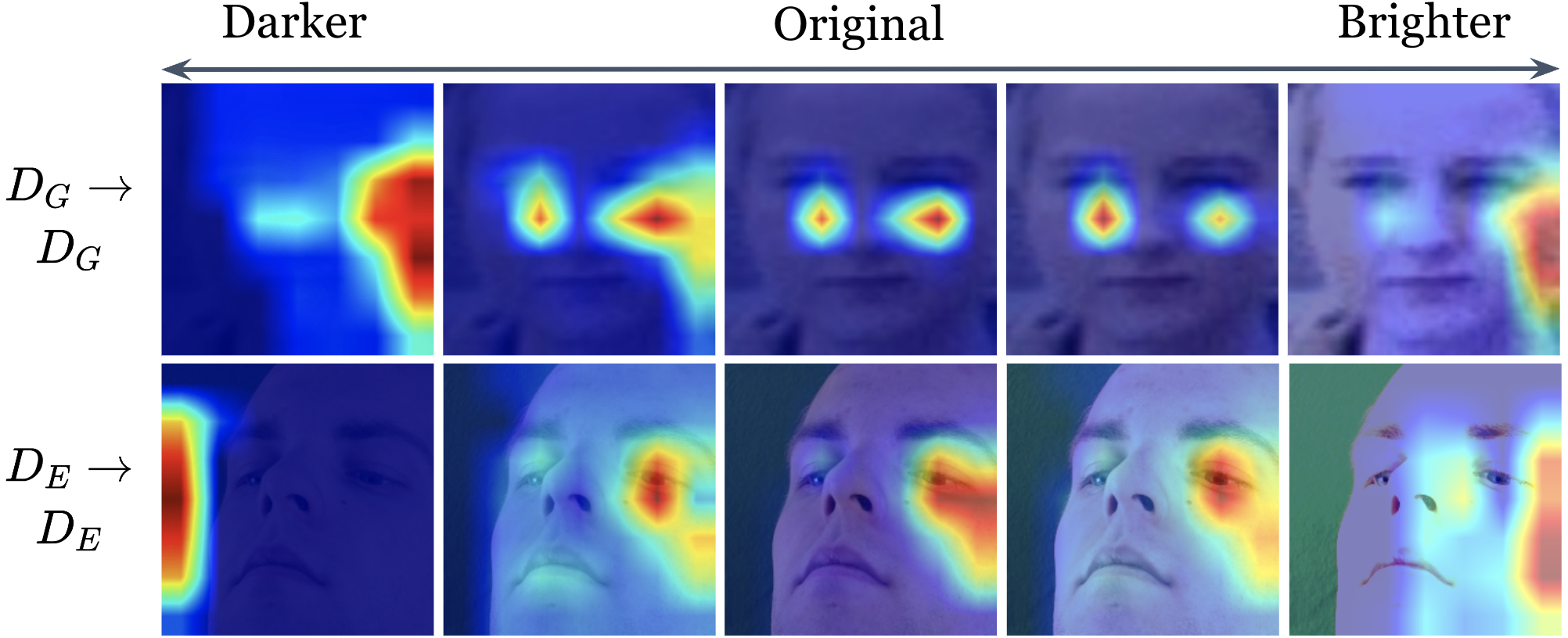}}%
\hfill
\subfloat[CauGE]{\label{fig:illum_cam_cauge} \includegraphics[width=0.49\linewidth]{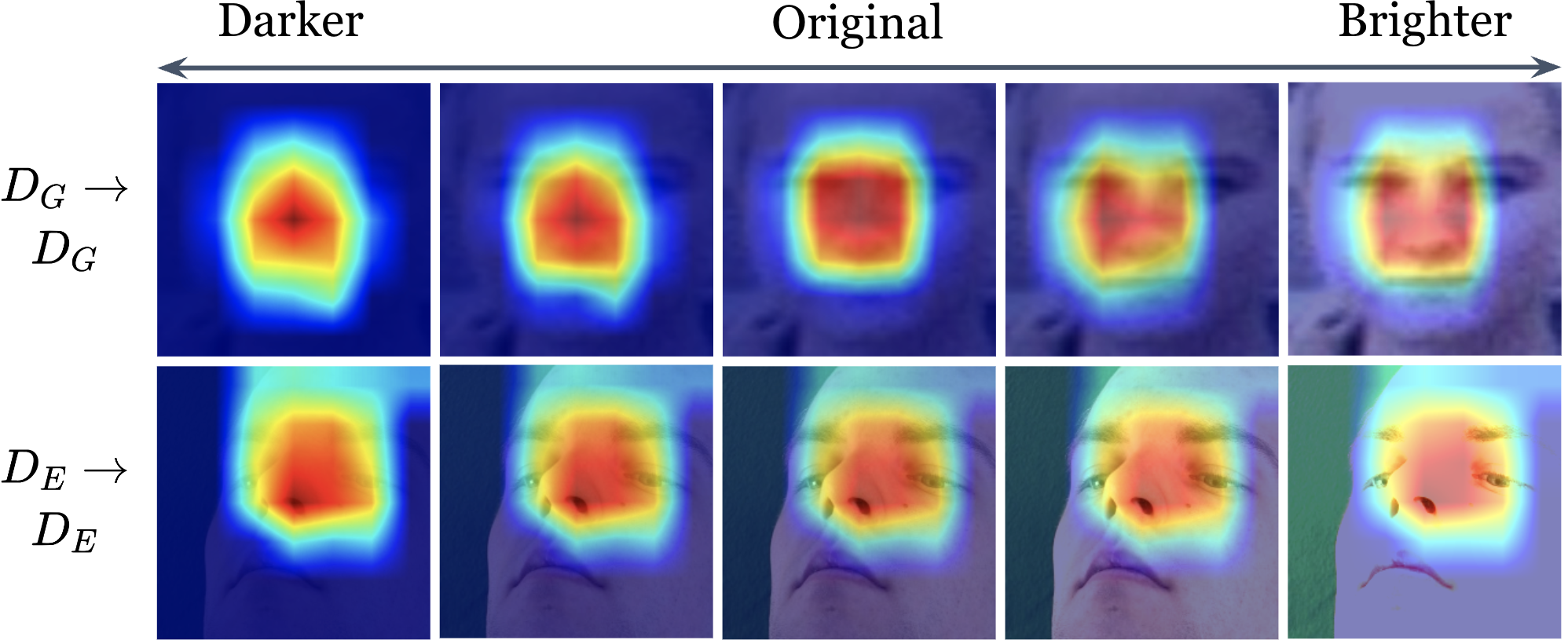}}%
\caption{Class Activation Map Visualization on varying illumination conditions. a) The baseline's class activation map is inconsistent with illumination change. b) CauGE's class activation map is consistent with illumination change.}
\label{fig:illum_gradcam}
\end{figure}

\begin{figure}[t!]
\centering
\subfloat[Baseline]{\label{fig:cross_cam_baseline} \includegraphics[width=0.48\linewidth]{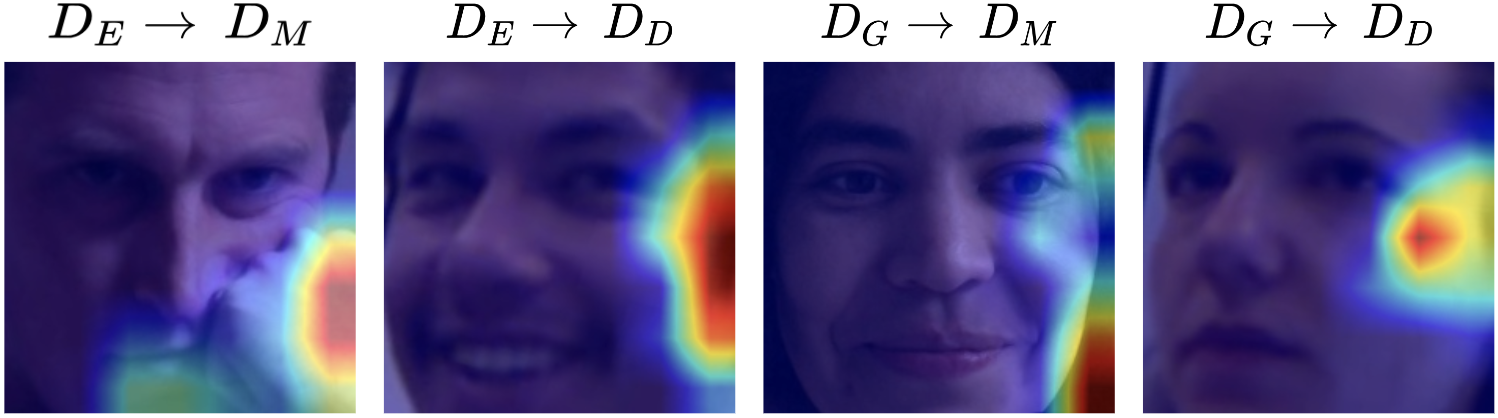}}%
\hfill 
\subfloat[CauGE]{\label{fig:cross_cam_cauge} \includegraphics[width=0.48\linewidth]{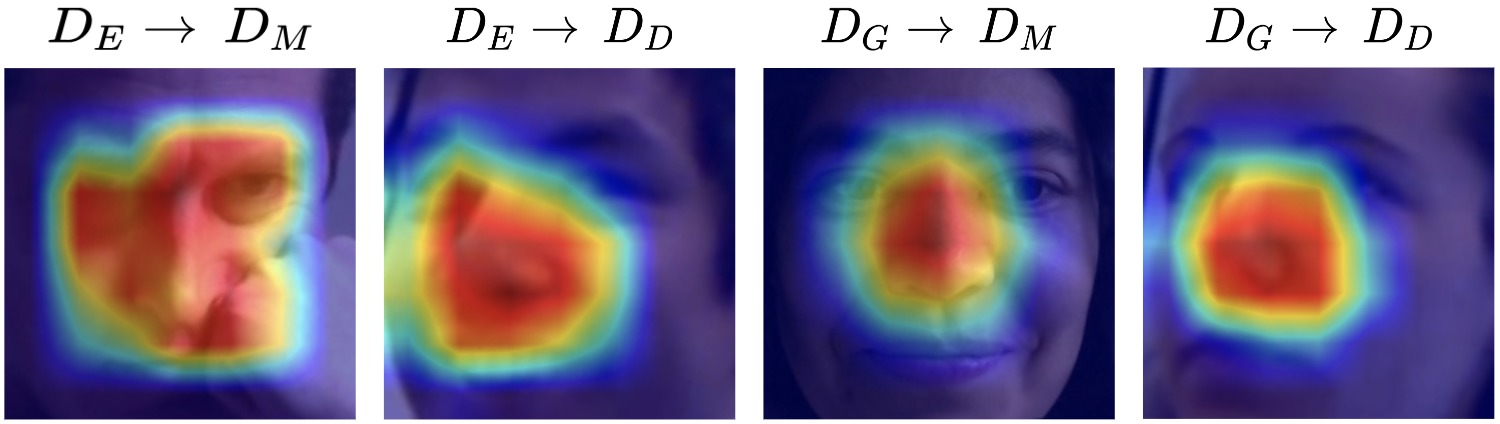}}%
\caption{Class Activation Map Visualization on cross-dataset condition. a) The baseline does not focus on the face region. b) CauGE focus on face region.}
\label{fig:cross_gradcam}
\end{figure}

\subsubsection{Class Activation Map Visualization}

In order to confirm that CauGE extracts features that meet the general principle of causal mechanisms, we have provided attention maps for the last convolutional layer of the feature extractor using a visualization technique proposed by Muhammad and Yeasin~\cite{muhammad_eigen-cam_2020}. To determine whether the model is consistent with domain shift, we randomly sample one identical image each from $D_E$ and $D_G$ and change illumination using implementation of Pillow Library\cite{murray_python-pillowpillow_2023}. As shown in Figure~\ref{fig:illum_gradcam}, the baseline's class activation map is inconsistent with illumination change, while CauGE's class activation map is consistent with illumination change. We interpret class activation map consistency as CauGE extracting domain-invariant features. Additionally, we show class activation map (See Figure~\ref{fig:cross_gradcam}) in the four tasks we previously conducted for experiments, $D_E \xrightarrow{} D_M$, $D_E \xrightarrow{} D_D$, $D_G \xrightarrow{} D_M$, and $D_G \xrightarrow{} D_D$. The non-face region mainly contributes to the baseline's output, while the face region contributes to CauGE's output.

\subsubsection{t-SNE Visualization}

\begin{figure}[t!]
\centering
\subfloat[Baseline]{\label{fig:tsne_baseline} \includegraphics[width=0.42\linewidth]{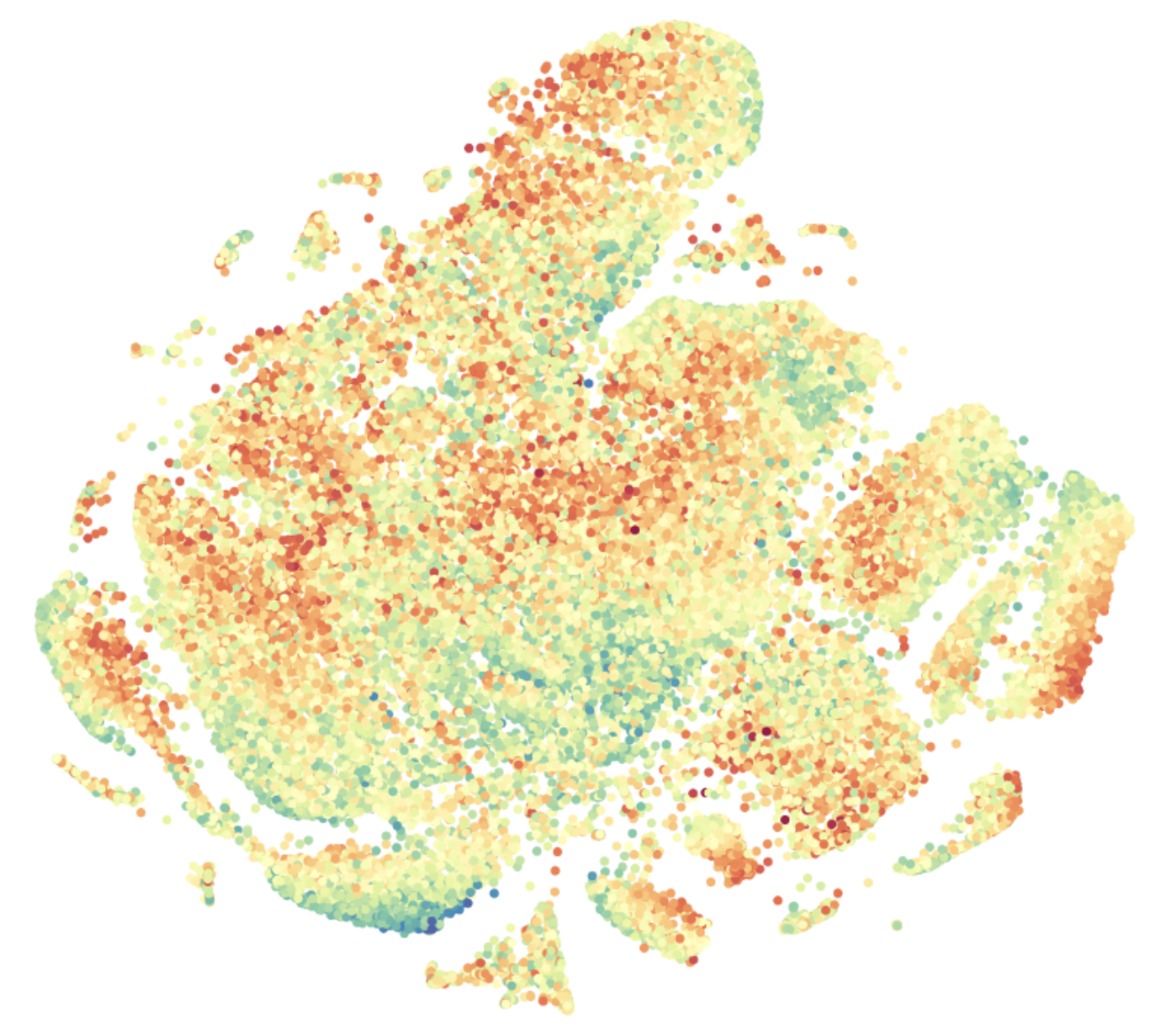}}
\hfill
\subfloat[CauGE]{\label{fig:tsne_cauge} \includegraphics[width=0.40\linewidth]{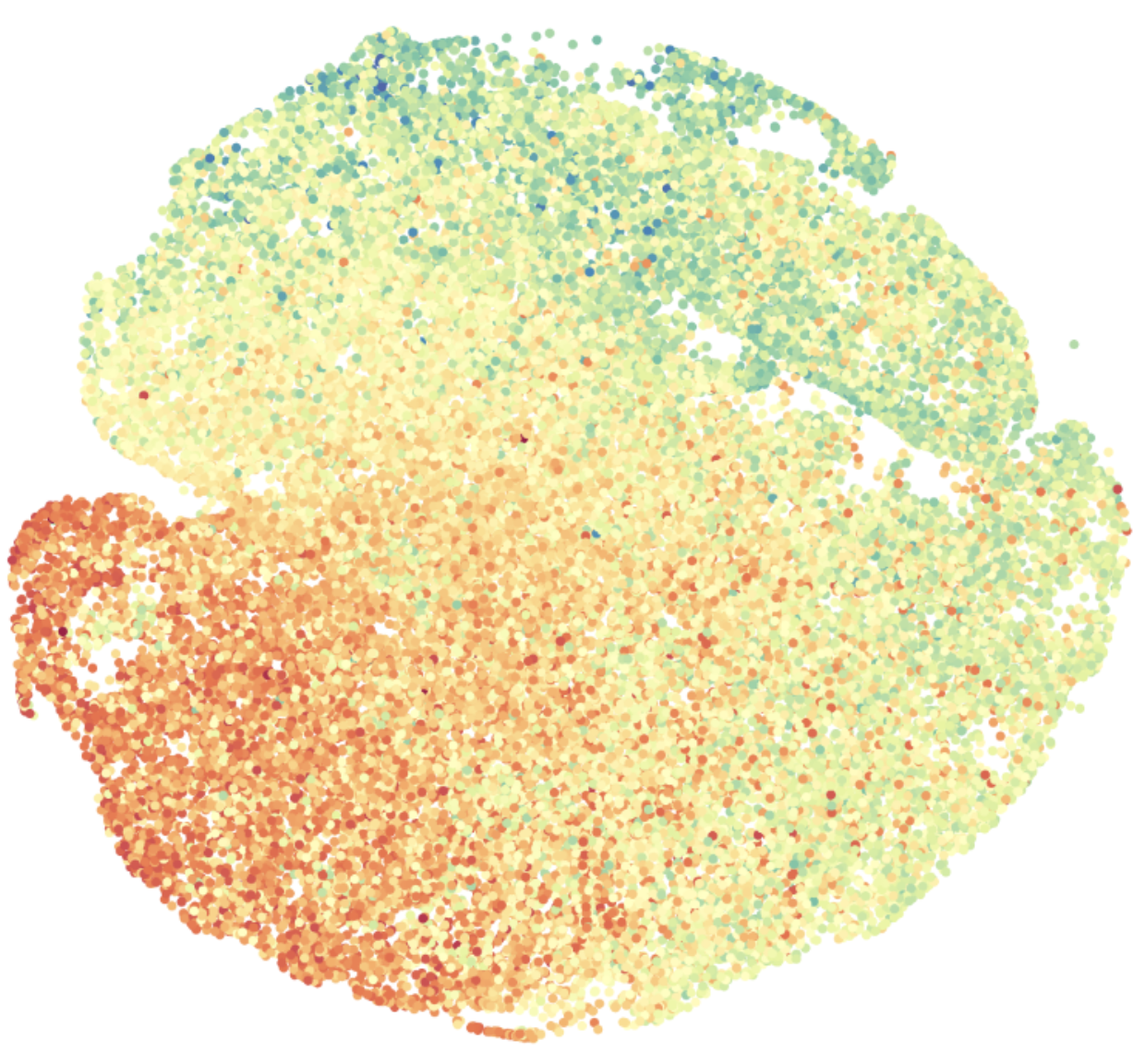}}%
\caption{t-SNE Visualization in $D_E \xrightarrow[]{} D_M$. Similar colors indicate similar gaze directions; a) Extracted features from the baseline are not correlated to gaze but are unsteady to domain shifts. b) Extracted features from CauGE are correlated to gaze and invariant to domain shifts.}
\label{fig:tsne}
\end{figure}

We have conducted a detailed analysis of the extraction of domain-invariant features. We utilize t-SNE~\cite{maaten_visualizing_2008} to display the $D_M$ gaze feature vectors obtained from both the baseline and CauGE, trained with $D_E$. As our primary task is gaze estimation, the extracted feature should be correlated to the gaze angle. Additionally, if the method extracts domain-invariant features, features should be associated to gaze unconditionally to the domain. The visualization results are shown in Figure~\ref{fig:tsne}. The t-SNE result by the baseline (See Figure~\ref{fig:tsne}(a)) is not correlated to gaze, as spurious correlation influences the model, which results from the model being domain-variant. In contrast, the t-SNE result by CauGE (See Figure~\ref{fig:tsne}(b)) correlates to gaze, as CauGE extracts causal factors of gaze angle, being invariant to the domain.

\section{Limitation}
\label{sec:limitation}
It is important to note the inherent limitations in the application of causal inference within the CauGE framework. Primarily, our approach utilizes causal mechanisms as tools to enhance model robustness and performance, focusing on empirical validation rather than exhaustive exploration of underlying causal relationships. This pragmatic approach emphasizes operational effectiveness over a complete causal discovery, given the complex interplay of biological, environmental, and perceptual factors that influence the human gaze. Consequently, while the causal models utilized within CauGE contribute to its efficacy, they may not fully capture the true causal dynamics of gaze behavior.

\section{Conclusion}
\label{sec:conclusion}
We present CauGE (Causal Representation-Based Domain Generalization on Gaze Estimation), a novel approach designed to tackle the complex task of domain generalization in gaze estimation. By incorporating the concept of causal representation, CauGE employs adversarial training and an additional penalizing term to achieve exceptional performance in generalizing gaze estimation across domains. Our experimental results demonstrate that CauGE displays remarkable accuracy and robustness, even in domain shifts and variations, highlighting its potential as a promising solution for real-world gaze estimation applications.


\bibliographystyle{unsrt}
\bibliography{ref}

\end{document}